\documentclass[12pt]{article}
\usepackage{openwork}

\usepackage{graphicx}
\usepackage{float} 
\usepackage{caption}
\usepackage{multirow}
\usepackage{multicol}
\usepackage{longtable}
\usepackage{booktabs}
\usepackage{array}
\usepackage{xcolor}
\usepackage{colortbl}

\usepackage{gb4e} 
\noautomath 

\title{Noisy memory encoding explains negative polarity illusions}
\author[1,2]{Yuhan Zhang}
\author[3]{Edward Gibson}
\affil[1]{Department of Linguistics, Stanford University (corresponding: yuhanczhang@gmail.com)}
\affil[2]{BIO-X Interdisciplinary Biosciences Institute, Stanford University}
\affil[3]{Department of Brain and Cognitive Sciences, Massachusetts Institute of Technology}
\date{}

\begin{document}
\maketitle

\vspace{0.3cm}

\begin{abstract}
A sentence like \textit{The authors that no critics recommended have ever received acknowledgment for a best-selling novel} is sometimes rated as acceptable even though, strictly speaking, it is ungrammatical because the negative polarity word \textit{ever} is not licensed where it is. This behavioral effect is sometimes called a ``negative polarity illusion''. Here we propose that the lossy context surprisal theory of Hahn et al. (2022) -- whereby people have an imperfect encoding of complex sentences -- might explain this effect. We hypothesize that people have poor memory representation of the determiners in the main-clause and embedded-clause subjects and could entertain a determiner exchange that licenses \textit{ever}. We propose that more similar determiners in those positions would trigger stronger illusion effects. Acceptability judgment tasks with six novel determiner pairs (e.g., \textit{few} and \textit{many}, \textit{few} and \textit{most}) support our proposal, showing, specifically, that a novel sentence, \textit{Many authors that few critics recommended have ever received acknowledgment for a best-selling novel}, triggered a much stronger illusion than the canonical one even without time pressure.
These results offer further support for the suggestion that human language processing is imperfect and resource-rational: in face of working memory limitations, humans rationally reconstruct what is most likely from noisy linguistic input to facilitate downstream processing.
\end{abstract}

\section{Introduction}

Language comprehension involves establishing syntactic dependencies between words. This process becomes effortful and error-prone when dependency distances are long, and especially when similar words occur between and before dependency connections \parencite{gibson_linguistic_1998, gibson_dependency_2001, gibson_dependency_2026, gordon_memory_2001,grodner_consequences_2005, just_capacity_1992,lewis_interference_1996,lewis_activationbased_2005,mcelree_memory_2003, van_dyke_retrieval_2006}. 
Such difficulty is extreme in the case of processing nested (or ``center-embedded'') structures, such as in the English relative clause example in (\ref{ex:centerembed_grammatical}) \parencite{frazier_syntactic_1985,miller1963finitary,miller_free_1964}. 

\begin{exe}
    \ex The apartment [that the maid [who the cleaning service sent over] cleaned] was well-decorated. \label{ex:centerembed_grammatical}
\end{exe}

Interestingly, sentences like (\ref{ex:centerembed_grammatical}) are so difficult to process that people often rate an ungrammatical version -- that lacks the second verb phrase \textit{cleaned} (\ref{ex:centerembed_illusion}) -- as equally acceptable \parencite{gibson1999memory}: 

\begin{exe}
    \ex * The apartment [that the maid [who the cleaning service sent over]  \underline{\textcolor{white}{clean}}] was well-decorated.\footnote{Linguists often annotate sentences like this with an asterisk (*) indicating that they are ungrammatical: not generated by the syntactic rules for the language.}\label{ex:centerembed_illusion}
\end{exe}

The acceptability of (\ref{ex:centerembed_illusion}) relative to (\ref{ex:centerembed_grammatical}) is sometimes referred to as a \textit{structural forgetting illusion}. This illusion was first observed in \textcite{frazier_syntactic_1985} as an intuition due to Janet Fodor.  \textcite{gibson1999memory} then provided acceptability judgment data from an experiment supporting the intuition. 

The best current explanation for the structural forgetting illusion is the lossy-context approach of \textcite{futrell_lossy-context_2020} and \textcite{hahn2022resource}. According to this approach, comprehenders have imperfect (``lossy'') memory representations for the language that they encounter. For an example like (\ref{ex:centerembed_grammatical}), a comprehender can't veridically maintain the initial input string, because of its complexity, and hence they forget less important words, which can be rationally reconstructed due to the statistics of the language or from the context. In practice, this means that some high-frequency function words are forgotten, and might be re-constructed later as more probable function words in the typical statistics of English. Here, the complementizer \textit{that} following the word \textit{apartment} might be reconstructed as a preposition like \textit{of}:

\begin{exe}
    \ex The apartment [\textbf{of} the maid [who the cleaning service sent over]] was well-decorated. \label{ex:centerembed_lossy}
\end{exe}

The sentence in (\ref{ex:centerembed_lossy}) is actually well-formed, with only one nested relative clause, and two verb phrases. If people sometimes rationally re-construct (\ref{ex:centerembed_grammatical}) as (\ref{ex:centerembed_lossy}), then they might rate (\ref{ex:centerembed_grammatical}) as grammatical. This word substitution can be viewed as an instantiation of rational noisy-channel inference in language processing \parencite{gibson_rational_2013, levy_noisy-channel_2008, shannon_mathematical_1948}.

In this paper, we examine another syntactic illusion: the \textit{negative polarity illusion} \parencite{drenhaus_processing_2005}. This illusion has to do with negative polarity expressions, such as \textit{ever}, \textit{any}, \textit{at all}, which only occur in certain linguistic environments such as negation and conditionals \parencite[][a.o.]{chierchia2004scalar, giannakidou2019negative, ladusaw1979polarity, linebarger1987negative}. Focusing on \textit{no} and \textit{ever} as the negative polarity licensor and licensee, \textit{ever} is only licensed if the clause that contains it is marked by \textit{no}, such as (\ref{npi_intro_correct}) in comparison to the ungrammatical (\ref{npi_intro_false}).

\begin{exe}
\ex \label{npi_intro}
\begin{xlist}
    \ex \textbf{No} student has \textbf{ever} been to Antarctica. \label{npi_intro_correct}
    \ex * \textbf{The} student has \textbf{ever} been to Antarctica. \label{npi_intro_false}
\end{xlist}
\end{exe}

The canonical sentence structure that gives rise to the negative polarity illusion is in the ungrammatical sentence (\ref{ex:nothe_illusion}) where \textit{ever} is supposed to occur in a negative linguistic environment but the main clause -- \textit{The authors have ever received acknowledgment} -- does not contain a negative word. There is a negative quantifier \textit{no} in the embedded relative clause but it does not structurally license \textit{ever}. A grammatical alternative is in (\ref{ex:nothe_grammatical}) where the sentence initial \textit{no} licenses \textit{ever}.

\begin{exe}
    \ex \label{ex:classic_nothe}
    \begin{xlist}
        \ex * The authors [that \textbf{no} critics recommended] have \textbf{ever} received acknowledgment for a best-selling novel. \label{ex:nothe_illusion}
        \ex \textbf{No} authors [that the critics recommended] have \textbf{ever} received acknowledgment for a best-selling novel. \label{ex:nothe_grammatical}
        \ex * The authors [that the critics recommended] have \textbf{ever} received acknowledgment for a best-selling novel. \label{ex:nothe_ungrammatical}
        \ex * The authors [that some critics \textbf{did not} recommend] have \textbf{ever} received acknowledgment for a best-selling novel. \label{ex:nothe_sententialnegation}
    \end{xlist}
\end{exe}

The illusion effect refers to the empirical finding that during a speeded acceptability judgment task, English speakers rate (\ref{ex:nothe_illusion}) to be more acceptable than its ungrammatical control (\ref{ex:nothe_ungrammatical}), although both should be ungrammatical according to English grammar \parencite{drenhaus_processing_2005,orth_negative_2021,parker_linguistic_2025,xiang_dependency-dependent_2013}. In eye-tracking and self-paced reading studies, the critical word \textit{ever} in the illusion prone condition (\ref{ex:nothe_illusion}) receives shorter reading time than the ungrammatical condition (\ref{ex:nothe_ungrammatical}) \parencite{vasishth_processing_2008,parker_negative_2016,xiang_dependency-dependent_2013}. In EEG studies, the critical word \textit{ever} in the illusion prone condition exhibited a reduced N400 effect compared to the ungrammatical condition (\ref{ex:nothe_ungrammatical}) and a P600 effect compared to the grammatical condition (\ref{ex:nothe_grammatical}) \parencite{drenhaus_processing_2005,xiang_illusory_2009}. In addition, only negative quantifiers in the embedded subject (e.g., \textit{no}, \textit{few}) have been reported to trigger an illusion effect; other types of licensors, such as the sentential negation \textit{not} and the negative adverb \textit{never} are immune to an illusion when embedded in a relative clause such as in (\ref{ex:nothe_sententialnegation}) \parencite{de-dios-flores_more_2016, muller_not_2019, orth_negative_2021}. Interestingly, the illusion in (\ref{ex:nothe_illusion}) disappears in a normal acceptability task, one which is not speeded \parencite{parker_negative_2016}. While existing theories attribute the illusion effect to failed memory retrieval or noisy pragmatic inference, none of them predicts the effect of licensor specificity and time pressure \parencite{deprez_negative_2020,orth_negative_2021,vasishth_processing_2005, vasishth_processing_2008,xiang_illusory_2009,xiang_dependency-dependent_2013} (see Discussion for detailed comments).

We propose here that \textcite{futrell_lossy-context_2020}'s and \textcite{hahn2022resource}'s lossy-context approach can be used to explain the negative polarity illusion. In particular, a sentence like (\ref{ex:nothe_illusion}) is complex given the object-extracted relative clause and the two structurally similar noun phrases, and so cannot be veridically retained in memory
People will not retain the most predictable items -- the function words -- and will rationally reconstruct them. Hence the determiners \textit{the} and \textit{no} might sometimes be misremembered and hence exchanged. When people encounter \textit{ever} and access the memory representation, they might retrieve this nonveridical representation which licenses \textit{ever}, resulting in the observed illusion. Furthermore, this hypothesis correctly predicts that the illusion will be stronger for tasks that involve less time to precisely encode the input, such as a speeded grammaticality task \parencite{parker_negative_2016}. It also correctly predicts the restriction to negative quantifiers as illusion triggering candidates because an exchange of \textit{the} and a negative quantifier is more likely than with the sentential negation \textit{not}, for instance.

This hypothesis makes an interesting prediction which we test here: the more similar the determiners are in the main-clause and embedded-clause subject position, the more likely they are to be misremembered in each other's position, leading to a higher chance of an illusion effect. This prediction is based on the observation that human working memory is subject to interference of similar words \parencite{baddeley2012working,crowder_principles_1976,lewis_interference_1996,lewis_computational_2006} so that similar determiners cause a stronger illusion.

Through six untimed acceptability experiments, we tested seven determiner pairs \{\textit{few}, \textit{many}\}, \{\textit{few}, \textit{most}\}, \{\textit{few}, \textit{all}\}, \{\textit{few}, \textit{the}\}, \{\textit{no}, \textit{many}\}, \{\textit{no}, \textit{all}\},  and \{\textit{no}, \textit{the}\} in the canonical illusion prone structure (\ref{ex:nothe_illusion}). We found evidence that sentences with more similar determiner pairs \{\textit{few}, \textit{many}\} and \{\textit{few}, \textit{most}\}\footnote{\textit{Few}, \textit{many}, and \textit{most} are the so-called vague or context-dependent quantifiers. The cardinality of the set of objects modified by these quantifiers is only clear via context (\textit{at least n} for \textit{many}, \textit{more than half} for \textit{most}, \textit{at most n} for \textit{few}) \parencite{bremnes2024interplay,partee1988many,pezzelle2018probing,szymanik_quantifiers_2016}. In comparison, the meanings of \textit{no} and \textit{the} are more transparent.} in (\ref{ex:fewmany_illusion_intro}) and (\ref{ex:fewmost_illusion_intro}) indeed give rise to stronger illusion effects than the canonical sentence with \{\textit{no}, \textit{the}\}(\ref{ex:nothe_illusion}). Furthermore, determiner similarity measured by word embedding cosine similarity scores across various computational models positively correlates with illusion strength, supporting the lossy-context memory encoding theory as a potential explanation for the negative polarity illusion.

\begin{exe}
    \ex  \label{ex:fewmany_fewmost_intro}
    \begin{xlist}
        \ex * \textbf{Many} authors [that \textbf{few} critics recommended] have \textbf{ever} received acknowledgment for a best-selling novel. \label{ex:fewmany_illusion_intro}
        \ex * \textbf{Most} authors [that \textbf{few} critics recommended] have \textbf{ever} received acknowledgment for a best-selling novel. \label{ex:fewmost_illusion_intro}   
    \end{xlist}
\end{exe}

\section{Methods}

\subsection{Determiner similarity}

We adopted the distributional hypothesis to represent determiner similarity -- words that share similar meanings or functions occur in similar distributions or contexts \parencite{harris_distributional_1954} -- and used the cosine distance between word embeddings of these determiners as the similarity score. We extracted word embeddings of the six determiners (\textit{the}, \textit{no}, \textit{few}, \textit{many}, \textit{most}, \textit{all}) from three types of word representation models -- GloVe \parencite{pennington_glove_2014}, fastText \parencite{bojanowski_enriching_2017}, BERT (from the last hidden layer) \parencite{devlin_bert_2019} -- to obtain an arithmetic average similarity score. We obtained the GloVe model from the open-source Python library \textit{gensim} \parencite{rehurek_lrec}. Its training corpora contains a combination of Wikipedia and Gigaword 5. The fastText model was also from \textit{gensim} and was trained on the Wikipedia data, UMBC webbase corpus and statmt.org news dataset. We obtained the BERT model from the open-source Python library \textit{transformers} \parencite{wolf-etal-2020-transformers}. We used the vanilla model variant (\texttt{bert-base-uncased}) that was trained on the BookCorpus and Wikipedia data. The dimensions of the word embedding vectors for GloVe, fastText, and BERT are 300, 300, and 768. We first calculated the pair-wise cosine similarity score among all determiners within one model and averaged the three scores per determiner pair. The aggregate score removes idiosyncrasies that might arise from models' training corpora, training regime, and algorithms.

\subsection{Offline grammaticality judgment tasks}

We conducted six experiments each consisting of an offline, untimed acceptability judgment task. Each experiment tested whether the illusion effect would occur given one of the critical determiner pairs from \{\textit{few}, \textit{many}\}, \{\textit{few}, \textit{most}\}, \{\textit{few}, \textit{all}\}, \{\textit{few}, \textit{the}\}, \{\textit{no}, \textit{many}\} and \{\textit{no}, \textit{all}\} in the sentence initial subject and the embedded subject. The control condition in each experiment was the same material from \textcite{parker_negative_2016}, featuring \{\textit{no}, \textit{the}\} as the determiner pair. The experimental and control conditions only differed by having different determiners. An example stimulus for the experiment of \{\textit{few}, \textit{many}\} is shown in Table \ref{tab:stimuli_few_many} (see Table \ref{tab:s1_stimuli_for_six} for example stimuli for all six experiments).

\begin{table}[ht]
\centering
\caption{Example stimuli for the experiment of \{\textit{few}, \textit{many}\}}
\vspace{0.2cm}
\begin{tblr}{
  width = \linewidth,
  colspec = {Q[l, wd=0.2\linewidth] Q[l, wd=0.52\linewidth] Q[l, wd=0.2\linewidth]},
  hlines,
  vlines,
  row{1-8} = {ht=1.2cm, m},
}
\SetCell[c=3]{l} The experimental condition & & \\
Grammatical    & \textbf{Few} authors that \textbf{many} critics recommended \ldots  & \SetCell[r=3]{m} \ldots\ have \textbf{ever} received acknowledgement for a best-selling novel. \\
Illusion-prone & \textbf{Many} authors that \textbf{few} critics recommended \ldots   & \\
Ungrammatical  & \textbf{Many} authors that \textbf{many} critics recommended \ldots & \\
\SetCell[c=3]{l} The control condition & & \\
Grammatical    & \textbf{No} authors that \textbf{the} critics recommended \ldots    & \SetCell[r=3]{m} \ldots\ have \textbf{ever} received acknowledgement for a best-selling novel. \\
Illusion-prone & \textbf{The} authors that \textbf{no} critics recommended \ldots    & \\
Ungrammatical  & \textbf{The} authors that \textbf{the} critics recommended \ldots   & \\
\end{tblr} \label{tab:stimuli_few_many}
\end{table}

Each experiment had 36 critical items, and each item had six conditions across the determiner pair and the grammaticality configuration (2 × 3 condition manipulation; the experiment with \{\textit{few}, \textit{the}\} had only five conditions because of the overlap of the ungrammatical condition between \{\textit{few}, \textit{the}\} and \{\textit{no}, \textit{the}\}). Besides the critical items, there were 72 filler items with relative clauses or complement clauses similar to the critical items but without the critical item \textit{ever}. An example sentence is \textit{No environmentalist’s conjecture that the local wildlife would be affected by the oil spill was discussed with the Coast Guard}. Within these fillers, 48 were grammatical and 24 were ungrammatical with errors such as subject-verb agreement mismatch and functional word deletion. Each stimulus was presented on the screen as a full sentence, accompanied by a YES/NO question (e.g., “Does this sentence mention critics who recommended authors?”). The answers were designed such that half were “yes” and half were “no”, as an attention checker. An acceptability question followed asking how natural the sentence is. Participants gave their judgment on a 7-point fully labeled Likert scale (1 = “Extremely unnatural”, 2 = “Unnatural”, 3 = “Somewhat unnatural”, 4 = “Neutral”, 5 = “Somewhat natural”, 6 = “Natural”, 7 = “Extremely natural”). Within each experiment, participants read a randomized list of 108 trials. The presentation of the trials followed a Latin Square design and the within-subjects design made sure that each participant would read the same number of trials across all the six conditions. Crucially, all six experiments gave participants ample time to provide their ratings. They could spend as long as needed for each trial.

\section{Data Availability}

All materials, data, and code are available via the OSF repository  \href{https://osf.io/4ywdc/overview?view_only=a6119b77b1854caaae1e3fa9cb819c4b}{https://osf.io/4ywdc}.

\section{Results}
\begin{figure}[ht]
    \centering
    \includegraphics[width=0.7\linewidth]{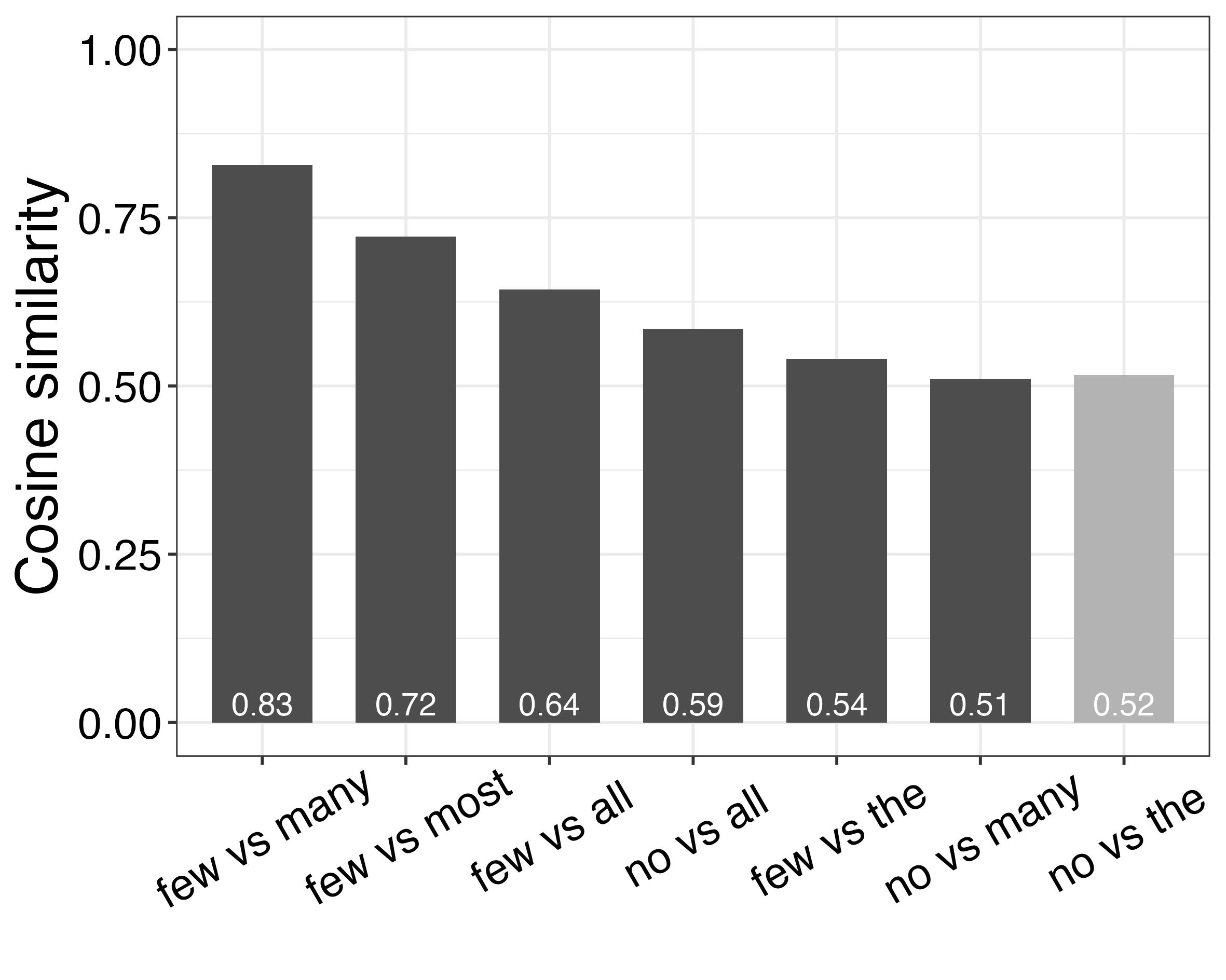}
    \caption{\textbf{Arithmetic mean of the cosine similarities across three embeddings} (i.e., GloVe, fastText, and BERT). The pair \{\textit{no}, \textit{the}\} is taken as the control comparison.}
    \label{fig:average_cosine_similarity}
\end{figure}

Figure \ref{fig:average_cosine_similarity} shows the averaged cosine similarity score for the critical determiner pairs that were tested in the acceptability judgment tasks. With the range between $[0,1]$, a higher cosine similarity indicates that the two determiners are more likely to occur in similar contexts and share similar meanings. \textit{Few} and \textit{many} are the most similar pairs ($0.83$), followed by \textit{few} and \textit{most} ($0.72$); the most dissimilar determiner pairs are \{\textit{no}, \textit{many}\} ($0.51$) and \{\textit{no}, \textit{the}\} ($0.52$) (See Figure \ref{fig:cosine_similarity_3models} for separate cosine similarity scores from all three embedding models.).

Around 50 participants were recruited from Prolific for each acceptability judgment experiment. We excluded data from those (a) who did not complete at least 90\% of trials; (b) who did not answer at least 75\% of the comprehension checks correctly; (c) who gave the same rating across all test trials; and/or (d) who self-identified as non-native speakers of English or from countries other than the United States. Across the experiments for \{\textit{few}, \textit{many}\}, \{\textit{few}, \textit{most}\}, \{\textit{few}, \textit{all}\}, \{\textit{few}, \textit{the}\}, \{\textit{no}, \textit{many}\} and \{\textit{no}, \textit{all}\}, we analyzed data from 48 (out of 51), 51 (out of 55), 51 (out of 55), 49 (out of 52), 45 (out of 48), and 46 (out of 49) participants respectively.

\begin{figure}[ht]
    \centering
    \includegraphics[width=\linewidth]{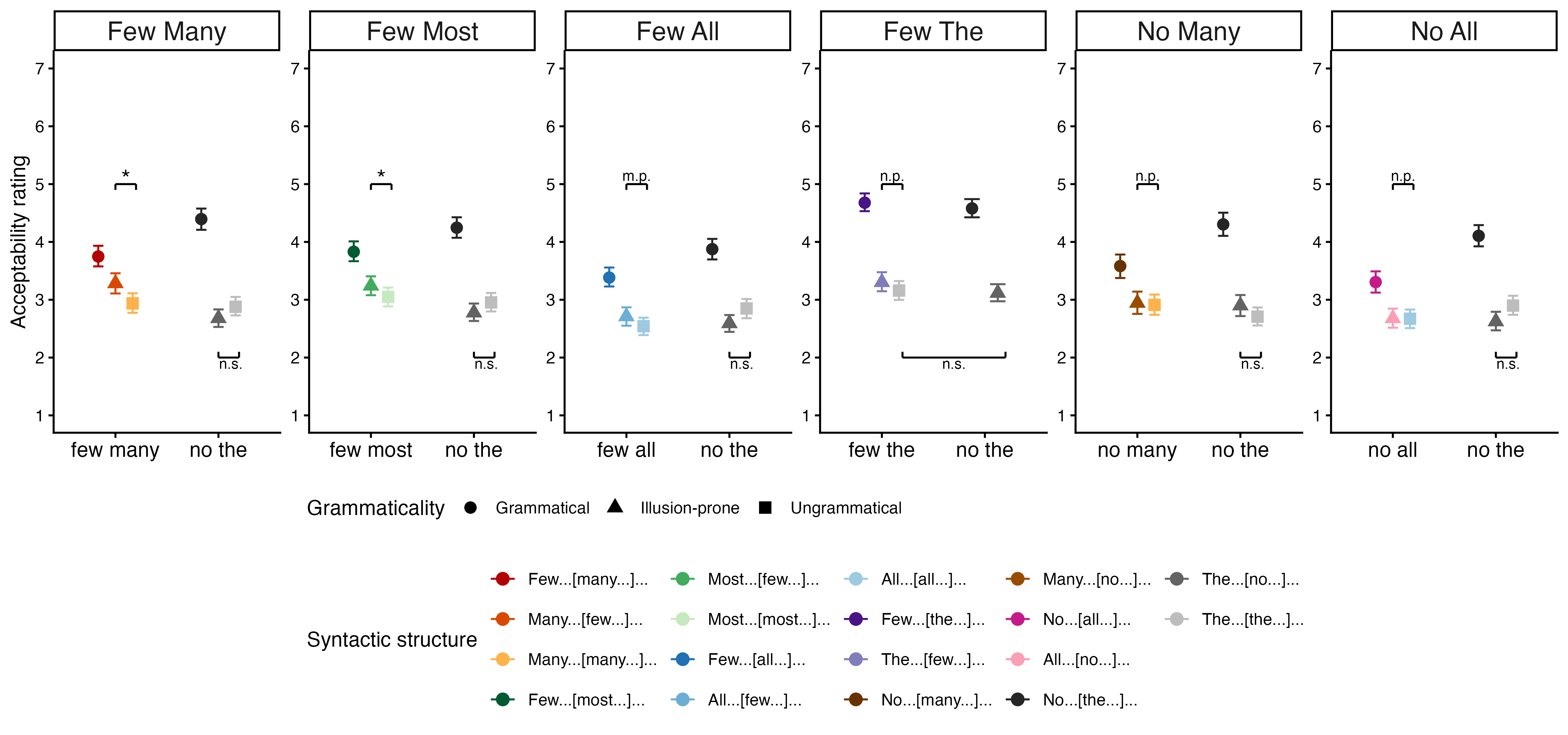}
    \caption{\textbf{Acceptability judgment ratings across six experiments.} Dots indicate average ratings with error bars representing 95\% bootstrapped confidence intervals. Dot colors indicate different syntactic structures. Dot shapes distinguish the grammaticality manipulation. \textbf{*} refers to positive ratings in the illusion-prone condition relative to the ungrammatical control condition, indicating an illusion effect. \textbf{m.p.} refers to marginally positive ratings and \textbf{n.p.} refers to the lack of positive ratings and thus an illusion effect. Statistical results are derived from Bayesian statistical modeling (see Table \ref{tab:posthoc_hypothesis_testing} for details).}
    \label{fig:six_acceptability_rating}
\end{figure}

Acceptability ratings for all six experiments are shown in Figure \ref{fig:six_acceptability_rating}. Sentences in nearly all conditions received neutral to somewhat unnatural ratings. We speculate that this was due to the intrinsic complexity of these sentences. Still, the variance among conditions is informative on the existence and strength of an illusion effect. We say that a particular pair of determiners exhibits an illusion effect when the illusion-prone condition received higher ratings than the ungrammatical condition, despite being technically ungrammatical. As shown in Figure \ref{fig:six_acceptability_rating}, \{\textit{few}, \textit{many}\} and \{\textit{few}, \textit{most}\} triggered a significant illusion effect while none of the baseline \{\textit{no}, \textit{the}\} conditions did. 

To quantify the illusion effect, we ran Bayesian multilevel cumulative ordinal regression models for each experiment separately. We used R (version 4.5.3) \parencite{R} and the package \textit{brms} \parencite{burkner2017brms, burkner2019ordinal} for modeling. In each model, the dependent variable was the raw rating from the Likert scale for each critical trial. The distance between intervals in the Likert scale were not assumed to equal. The determiner manipulation and the grammaticality manipulation were entered as 2 x 3 dummy-coded fixed effects. The reference levels were always the determiner control of \{\textit{no}, \textit{the}\} and the ungrammatical condition. We added an interaction term between the main effects because the theoretical interest is that different determiner pairs might cause differential illusion effects. To capture the maximal complexity recommended for mixed-effects models \parencite{barr2013random}, we included random intercepts and slopes for the full fixed-effects structure for both subjects and items\footnote{The model structure is brm(rating $\sim$ determiner * grammaticality + (determiner * grammaticality$|$participant) + (determiner * grammaticality$|$item), family = cumulative(link=``probit'', threshold=``flexible''), iter=4000, chains=4)}. The model structure for the \{\textit{few}, \textit{the}\} experiment was different. Because both determiner pairs had the ungrammatical condition feature the determiner pair \{\textit{the}, \textit{the}\}, we coded this ungrammatical control as the reference level and the other four conditions as four independent levels. In this way, the illusion effect for both determiner pairs would be manifested as main effects. Similar to the other models, the random effect here included an intercept and a slope for the condition for both subjects and items. Across all models, the prior distributions for all the intercepts and coefficients of fixed effects were set as a normal distribution with a mean of 0 and a standard deviation of 2 (i.e., Normal(0,2)); the prior for the correlation matrices of random effects was set to be LKJ(2) where LKJ is the default weakly informative prior for correlation matrices in \textit{brms} \parencite[e.g.,][]{lewandowski2009generating}. For the other parameters, we chose the default priors in \textit{brms}. Each model had four parallel sampling chains, each with 4000 iterations and the first 2000 as a warmup. All $\hat{R}$s for group-level parameters were under 1.01, suggesting that the sample chains had converged satisfactorily to the target posterior distribution. See Tables \ref{tab:s_fewmany_model_result} to \ref{tab:s_noall_model_stats} for statistical summaries of each model. 

We estimated the illusion effect using directional posterior contrasts derived from the fitted Bayesian cumulative ordinal models. Specifically, using the hypothesis() function in \textit{brms}, we constructed linear combinations of the fixed-effect parameters corresponding to the contrast between the illusion-prone condition and the ungrammatical condition for each determiner pair. A positive value of this contrast indicates that sentences in the illusion-prone condition were estimated to receive higher acceptability ratings than their ungrammatical counterparts, which we interpret as evidence for an illusion effect.\footnote{For example, in the {\textit{few}, \textit{many}} experiment, the contrast for the {\textit{few}, \textit{many}} determiner pair was specified as ``grammaticality-illusion + determiner-few...many:grammaticality-illusion $> 0$''. The corresponding contrast for the reference determiner pair \{\textit{no}, \textit{the}\} was specified as ``grammaticality-illusion $> 0$''.} For each contrast, we summarized the full posterior distribution of the relevant linear combination by reporting its posterior mean, 95\% credible interval, evidence ratio, and the posterior probability that the contrast was greater than zero (see Table~\ref{tab:posthoc_hypothesis_testing}). For the one-sided hypotheses considered here, the evidence ratio computed by \textit{brms} corresponds to the posterior probability of the contrast being greater than zero divided by the posterior probability of the contrast being less than zero. Thus, evidence ratios larger than one indicate that the posterior distribution assigns more probability to a positive illusion effect than to its directional alternative. We report posterior probabilities directly rather than treating them as frequentist $p$-values. As a conservative descriptive criterion, we mark contrasts for which the posterior probability of a positive effect exceeds 0.975, corresponding to a posterior odds ratio greater than 39:1 in favor of a positive contrast. The linear combinations for the experiment with \{\textit{few}, \textit{the}\} are just two of their main effects.

\begin{table}[ht]
\caption{Hypothesis testing for an illusion effect across six experiments} \label{tab:posthoc_hypothesis_testing}
\centering
\begin{tabular}{p{0.8cm}p{2cm}p{1.7cm}p{1.5cm}p{1.5cm}p{1.5cm}p{1.5cm}p{1.5cm}}
\hline
\textbf{Exp} & \textbf{DET} & \textbf{Post.mean} & \textbf{Est.Error} & \textbf{CI.lower} & \textbf{CI.upper} & \textbf{Evi.Ratio} & \textbf{Post.Prob} \\
\hline
\multirow{2}{*}{1} & \textit{few} \& \textit{many} & 0.39 & 0.12 & 0.15 & 0.63 & 665.67 & 1* \\
                   & \textit{no} \& \textit{the}   & -0.19 & 0.11 &-0.41 & 0.03 & 0.05  & 0.05 \\
\hline
\multirow{2}{*}{2} & \textit{few} \& \textit{most} & 0.24 & 0.12 & 0.01 & 0.46 & 50.28 & 0.98*  \\
                   & \textit{no} \& \textit{the}  & -0.22 & 0.10 & -0.41 & -0.02 & 0.01 & 0.01  \\
\hline
\multirow{2}{*}{3} & \textit{few} \& \textit{all} & 0.21 & 0.11 & -0.01 & 0.43 & 31.92 & 0.97  \\
                   & \textit{no} \& \textit{the}  & -0.26 & 0.10 & -0.45 & -0.07 & 0 & 0  \\
\hline
\multirow{2}{*}{4} & \textit{few} \& \textit{the} & 0.12 & 0.12 & -0.11 & 0.35 & 5.44 & 0.84  \\
                   & \textit{no} \& \textit{the}  & -0.06 & 0.11 & -0.27 & 0.15 & 0.40 & 0.29  \\
\hline
\multirow{2}{*}{5} & \textit{no} \& \textit{many} & 0.01 & 0.12 & -0.22 & 0.25 & 1.22 & 0.55  \\
                   & \textit{no} \& \textit{the}  & 0.11 & 0.12 & -0.12 & 0.33 & 4.59 & 0.82  \\
\hline
\multirow{2}{*}{6} & \textit{no} \& \textit{all}  & 0.02 & 0.11 & -0.19 & 0.23 & 1.40 & 0.58  \\
                   & \textit{no} \& \textit{the}  & -0.30 & 0.10 & -0.49 & -0.11 & 0 & 0  \\
\hline
\multicolumn{8}{p{14cm}}{\textit{Note: * means marked contrasts for which the posterior probability of a positive effect exceeds 0.975, indicating the existence of an illusion effect.}} \\
\hline
\end{tabular}
\end{table}

\begin{figure}[ht!]
    \centering
    \includegraphics[width=0.8\linewidth]{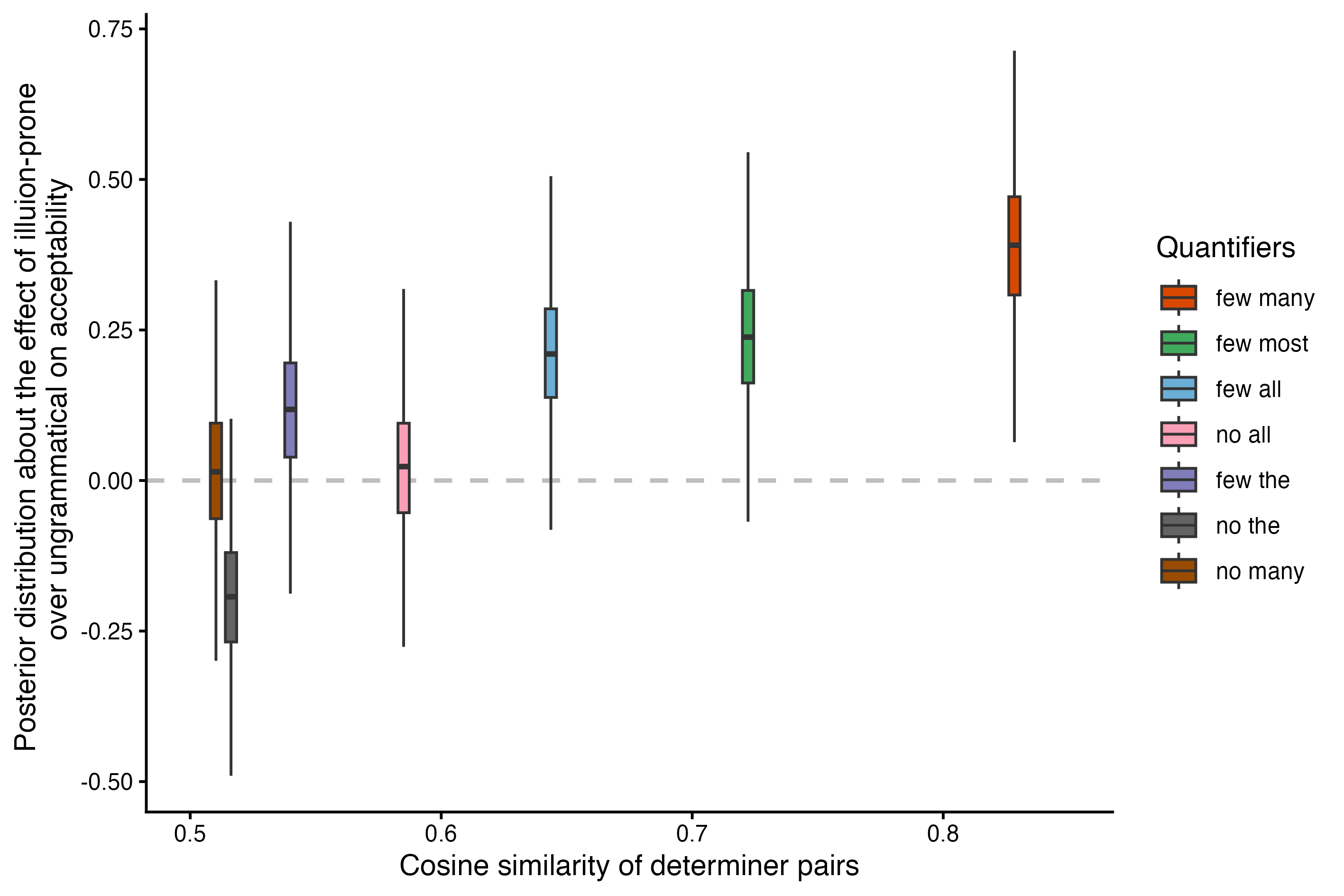}
    \caption{\textbf{Descriptive correlation between cosine similarity of determiner pairs and posterior illusion strength between the illusion-prone condition and the ungrammatical control condition.} Boxplots represent the distribution of the linear combination results from the posterior contrasts of each Bayesian ordinal regression models. The short horizontal bars represent the median in the distribution. The higher the distribution and the further away it is from zero, the stronger the illusion effect.  The \{\textit{no}, \textit{the}\} condition comes from the experiment of \{\textit{few}, \textit{many}\}. See Fig. \ref{fig:nothe_6exps} for the \{\textit{no}, \textit{the}\} condition in all six experiments.}
    \label{fig:correlation_acceptability_cosine}
\end{figure}

In sum, across all six experiments, only the determiner pairs of \{\textit{few}, \textit{many}\} and \{\textit{few}, \textit{most}\} give rise to an illusion effect. None of the control group with \{\textit{no}, \textit{the}\} shows an illusion effect, replicating \textcite{parker_negative_2016}. Figure \ref{fig:correlation_acceptability_cosine} shows a positive relationship between the cosine similarity of determiner pairs and their associated illusion strength estimated from behavioral experiments. Because the number of determiner pairs was small and posterior draws within each determiner pair are not independent observations, we did not conduct a formal correlation test. But with a descriptive Pearson correlation of $0.877$, this trend supports our hypothesis that the more similar determiners are, the stronger the illusion will be in the structure of (\ref{ex:nothe_illusion}).

\section{Discussion} \label{sec:Discussion}


Negative polarity illusions refer to sentences such as (\ref{ex:nothe_illusion}) where the word \textit{ever} does not appear in a negative polarity licensing environment but the sentence is rated more natural than its ungrammatical control (\ref{ex:nothe_ungrammatical}) \parencite[][a.o.]{drenhaus_processing_2005, parker_negative_2016, vasishth_processing_2008,xiang_illusory_2009}. 
We show that the lossy-context theory of language processing proposed by \textcite{futrell_lossy-context_2020} and \textcite{hahn2022resource} has the potential to provide an explanation.
When the linguistic input is complex, comprehenders might store a lossy memory representation, forgetting the most predictable, high-frequency function words, and reconstructing them according to the context and the statistics of the language.
The naturalness in (\ref{ex:nothe_illusion}) could result from a nonveridical memory representation of the context prior to \textit{ever} where the determiners in the main-clause and embedded-clause subjects are exchanged (e.g., \textit{the} in \textit{the authors} and \textit{no} in \textit{no critics}) due to their high predictability and their positional proximity. This imperfect representation licenses \textit{ever} and causes an illusion. 
This account predicts that time pressure in processing boosts the illusion strength as shorter time causes more imprecise memory encoding. It also predicts that the illusion is restricted to negative quantifiers such as \textit{no} and \textit{few} because word exchanges are only possible between the sentence initial \textit{the} and an embedded negative quantifier, not other types of negative polarity licensors such as \textit{not} or \textit{never} (\ref{ex:nothe_sententialnegation}). Both predictions are supported by existing empirical findings: (a) an illusion from (\ref{ex:nothe_illusion}) appears only in speeded acceptability tasks and disappears with unbounded time for judgment; (b) an illusion only appears with negative quantifiers as the potential licensor and disappears with other types of licensors \parencite{de-dios-flores_more_2016,muller_not_2019,orth_negative_2021,parker_negative_2016}.

In this paper, we tested a novel prediction of the lossy-context theory: the more similar the determiners are in the main-clause and embedded-clause subject positions, the stronger the illusion would be. Our investigation involved six untimed acceptability judgment experiments each of which tested the illusion strength associated with a novel determiner pair and compared that to the canonical pair \{\textit{no}, \textit{the}\} in \textcite{parker_negative_2016}. 
We found that determiner similarity as measured by cosine distance from language models positively correlated with the illusion strength (Fig.\ref{fig:correlation_acceptability_cosine}); the more similar pairs \{\textit{few}, \textit{many}\} and \{\textit{few}, \textit{most}\} triggered a strong illusion in sentences like (\ref{ex:fewmany_illusion_intro}) and (\ref{ex:fewmost_illusion_intro}) even without time restriction. These findings support the lossy memory theory of the illusion. 

A related theory of \textcite{vasishth_processing_2008} attributes the illusion to inaccurate retrieval from a veridical memory representation, rather than imperfect encoding or maintenance in working memory \parencite{lewis_activationbased_2005,lewis_computational_2006,vasishth_processing_2005}. This is caused by partial feature overlap between \textit{the} and \textit{no}. When a comprehender encounters the negative polarity word \textit{ever} in  (\ref{ex:nothe_illusion}), she will look for a potential licensor in working memory to form a dependency. She checks in parallel whether any preceding word is a word of negation and marks a clause that contains \textit{ever}. When she encounters \textit{no} whose negation feature partially fulfills the licensing requirement, she could falsely establish dependency with \textit{ever} in a probabilistic manner. This retrieval theory does not offer a ready explanation for the licensor specificity issue. Later work in \textcite{parker_negative_2016} shows that this proposal fails to simulate the reading behavior in English when the dependency gets longer with more intervening words. Therefore, this theory would need further work to capture the empirical landscape.

A further alternative theory resorts to contrastive inference in pragmatic processing as its explanation \parencite{mendia_spurious_2018, xiang_illusory_2009, xiang_dependency-dependent_2013}. These researchers propose that a sentence like \textit{The authors that no critics recommended have ever received acknowledgment} in (\ref{ex:nothe_illusion}) generates a contrastive inference such as \textit{The authors that \textbf{some} critics recommended have \textbf{not} ever received acknowledgment}, which licenses \textit{ever}. This inference is similar to the inference where \textit{The red apples are sweet} implies that \textit{Apples that are \textbf{not} red are \textbf{not} sweet} \parencite[e.g.,][]{altmann_interaction_1988, tanenhaus1995integration}. But, this account incorrectly predicts that all types of negative polarity licensors would trigger a contrastive inference leading to an illusion and thus overgeneralizes as well.

A third account from the literature rightly points out that the illusion effect is only restricted to negative quantifiers as the licensors based on principled acceptability judgment experiments on a wide range of licensors \parencite{orth_negative_2021, orth2023active, orth2025positive}. This account proposes that comprehenders undergo a mental process that transforms the relative clause to a syntactically higher position where \textit{no} adequately licenses \textit{ever}. This transformed hierarchical representation is linearly represented as [\textit{That no critics recommended}] \textit{the authors have ever received acknowledgment}. However, this account remains incomplete unless the proposed syntactic transformation can be independently motivated as a general mental operation in sentence comprehension.

Our proposal aligns with \textcite{parker_negative_2016} in arguing that the encoding mechanism is the source of illusion. They claim that the encoding of the preceding context changes over time. They empirically show that the canonical illusion effect in (\ref{ex:nothe_illusion}) disappears in a speeded judgment task when the dependency distance gets longer. An example sentence is *\textit{The journalists} [\textit{that \textbf{no} editors recommended for the assignment}] \textit{thought that the readers would \textbf{ever} understand the complicated situation}. 
The lossy-context approach readily predicts this observation. As activations of early linguistic input decay over time and the representation becomes lossy as more words are integrated, comprehenders could just forget the details of the input and fail to recall any potential negative polarity licensor in the representation. It is thus easier to detect an unlicensed \textit{ever}.


Overall, the lossy-context theory \parencite{futrell_lossy-context_2020,hahn2022resource} has an advantage over other theories in explaining the empirical observations and making novel predictions of negative polarity illusion. Its essence is that even under memory constraints and imperfect representation, comprehenders can still leverage the knowledge of language and the information from the context to rationally reconstruct a likely representation and continue the downstream processing. Underlying this resource-rational theory is the more general noisy-channel framework of sentence processing \parencite{gibson_rational_2013,levy_noisy-channel_2008,shannon_mathematical_1948} which has offered potential explanations for diverse language illusions and processing phenomena across the field \parencite[][a.o.]{chen_effect_2023,poliak_it_2023, ryskin_erp_2021,zhang_comparative_2025,zhang2023noisy,zhang2025graded}. Taken together, the present findings suggest that negative polarity illusion is not an isolated grammatical anomaly, but a revealing case study of how rational inference operates over noisy linguistic representations.

\section{Declaration of generative AI and AI-assisted technologies in the \\ manuscript preparation process}

During the preparation of this work the authors used Claude Sonnet 4.6 in order to generate basic R code to finish statistical analyses and visualization. After using this tool, the authors reviewed and edited the code as needed and take full responsibility for the content of the published article.

\section{Acknowledgements}

We thank Moshe Poliak, members in Ted's lab, and audiences at the 37th Annual Conference on Human Sentence Processing in 2024 for their helpful feedback. The research is funded by a National Science Foundation grant (BCS-2121074). Y.Z. is supported by a Stanford Bio-X Postdoctoral Fellowship.

\vspace{1em}

\printbibliography
\newpage

\appendix
\section{Appendix}
\renewcommand{\thetable}{S\arabic{table}}
\renewcommand{\thefigure}{S\arabic{figure}}
\setcounter{table}{0}
\setcounter{figure}{0}
\renewcommand{\thesection}{\Alph{section}}
\setcounter{section}{0}
\begin{figure}[ht]
    \centering
    \includegraphics[width=0.8\linewidth]{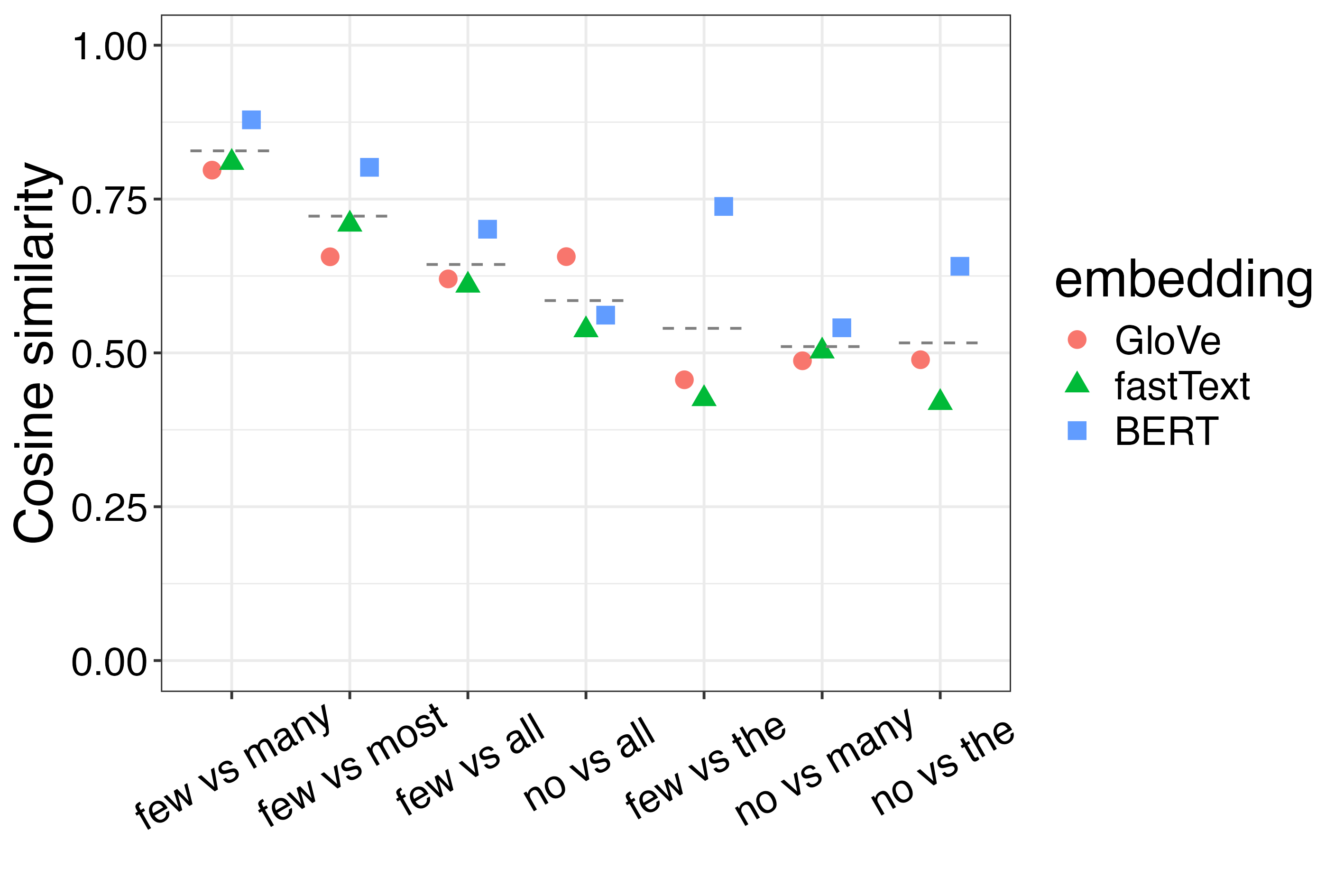}
    \caption{\textbf{Cosine similarity for seven pairs of determiners across three word embedding models} (dashed gray lines represent the average mean in Figure 1).}
    \label{fig:cosine_similarity_3models}
\end{figure}

\begin{figure}
    \centering
    \includegraphics[width=\linewidth]{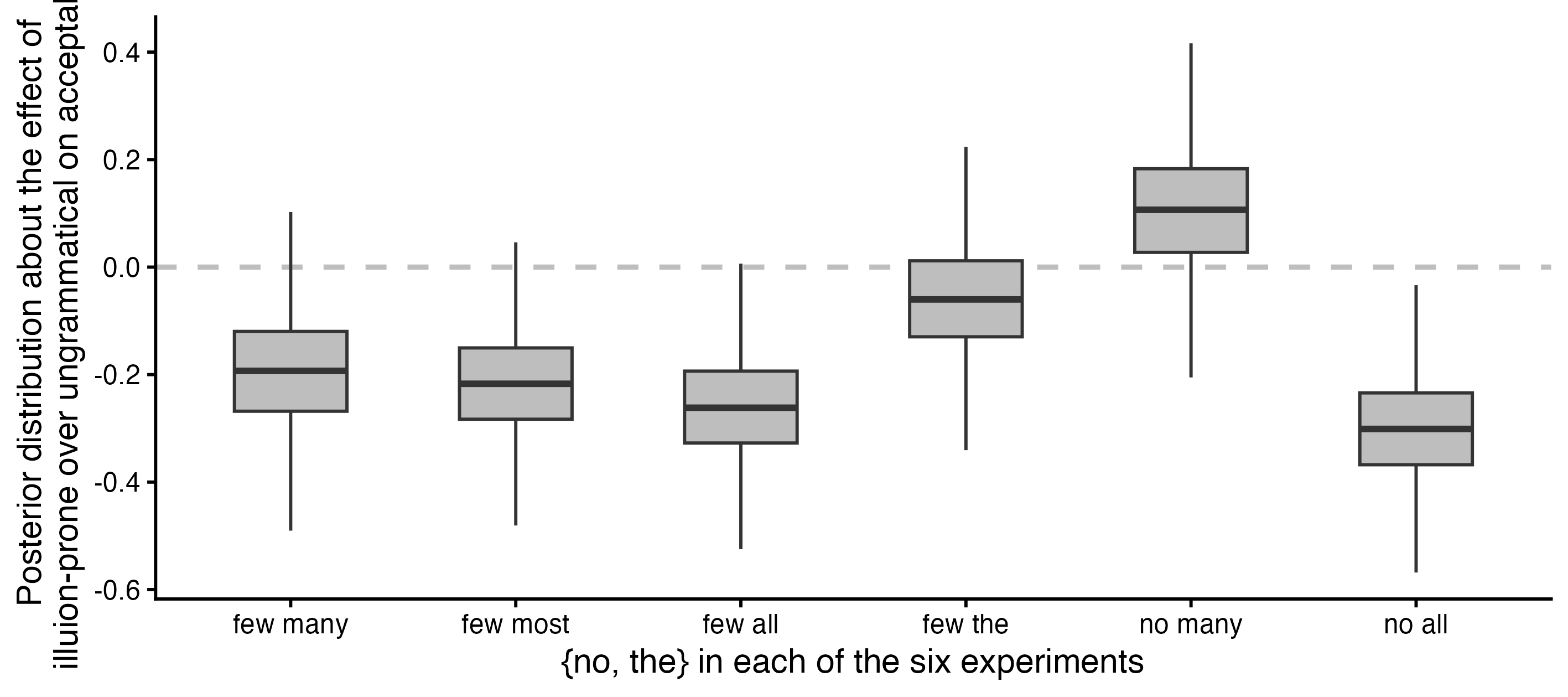}
    \caption{\textbf{Posterior acceptability difference between the illusion-prone condition and the ungrammatical condition for the determiner pair \{\textit{no}, \textit{the}\} across all six experiments.} Boxplots represent the distribution of posterior draws from each Bayesian ordinal regression models comparing the mean of acceptability between the two conditions. The fact that $y > 0$ indicates a significant illusion effect in that determiner pair.}
    \label{fig:nothe_6exps}
\end{figure}

\newpage
\begin{longtblr}[
  caption = {The example stimuli for all six offline acceptability judgment experiments},
  label = {tab:s1_stimuli_for_six},
]{
  width = \linewidth,
  colspec = {Q[l, wd=0.2\linewidth] Q[l, wd=0.52\linewidth] Q[l, wd=0.23\linewidth]},
  hlines,
  vlines,
  row{1,5,9,13,17,21,25,29} = {font=\normalfont},
}

\SetCell[c=3]{l} The experimental condition for \{\textit{few, many}\} & & \\
Grammatical    & \textbf{Few} authors that \textbf{many} critics recommended \ldots  & \SetCell[r=3]{m} \ldots\ have \textbf{ever} received acknowledgement for a best-selling novel. \\
Illusion-prone & \textbf{Many} authors that \textbf{few} critics recommended \ldots   & \\
Ungrammatical  & \textbf{Many} authors that \textbf{many} critics recommended \ldots & \\

\SetCell[c=3]{l} The experimental condition for \{\textit{few, most}\} & & \\
Grammatical    & \textbf{Few} authors that \textbf{most} critics recommended \ldots  & \SetCell[r=3]{m} \ldots\ have \textbf{ever} received acknowledgement for a best-selling novel. \\
Illusion-prone & \textbf{Most} authors that \textbf{few} critics recommended \ldots   & \\
Ungrammatical  & \textbf{Most} authors that \textbf{most} critics recommended \ldots & \\

\SetCell[c=3]{l} The experimental condition for \{\textit{few, all}\} & & \\
Grammatical    & \textbf{Few} authors that \textbf{all} critics recommended \ldots  & \SetCell[r=3]{m} \ldots\ have \textbf{ever} received acknowledgement for a best-selling novel. \\
Illusion-prone & \textbf{All} authors that \textbf{few} critics recommended \ldots   & \\
Ungrammatical  & \textbf{All} authors that \textbf{all} critics recommended \ldots & \\

\SetCell[c=3]{l} The experimental condition for \{\textit{few, the}\} & & \\
Grammatical    & \textbf{Few} authors that \textbf{the} critics recommended \ldots  & \SetCell[r=3]{m} \ldots\ have \textbf{ever} received acknowledgement for a best-selling novel. \\
Illusion-prone & \textbf{The} authors that \textbf{few} critics recommended \ldots   & \\
Ungrammatical  & \textbf{The} authors that \textbf{the} critics recommended \ldots & \\

\SetCell[c=3]{l} The experimental condition for \{\textit{no, many}\} & & \\
Grammatical    & \textbf{No} authors that \textbf{many} critics recommended \ldots  & \SetCell[r=3]{m} \ldots\ have \textbf{ever} received acknowledgement for a best-selling novel. \\
Illusion-prone & \textbf{Many} authors that \textbf{no} critics recommended \ldots   & \\
Ungrammatical  & \textbf{Many} authors that \textbf{many} critics recommended \ldots & \\

\SetCell[c=3]{l} The experimental condition for \{\textit{no, all}\} & & \\
Grammatical    & \textbf{No} authors that \textbf{all} critics recommended \ldots  & \\
Illusion-prone & \textbf{All} authors that \textbf{no} critics recommended \ldots   & \SetCell[r=2]{m} \ldots\ have \textbf{ever} received acknowledgement for a best-selling novel. \\
Ungrammatical  & \textbf{All} authors that \textbf{all} critics recommended \ldots & \\

\SetCell[c=3]{l} The control condition for all six experiments & & \\
Grammatical    & \textbf{No} authors that \textbf{the} critics recommended \ldots  & \SetCell[r=3]{m} \ldots\ have \textbf{ever} received acknowledgement for a best-selling novel. \\
Illusion-prone & \textbf{The} authors that \textbf{no} critics recommended \ldots   & \\
Ungrammatical  & \textbf{The} authors that \textbf{the} critics recommended \ldots & \\

\end{longtblr}

\begin{table}[ht]
\caption{Statistical results for the model in Experiment 1 (\textit{few}, \textit{many})}
\centering
\begin{tabular}{p{3.7cm}lllllll}
\hline
\textbf{Variable} & \textbf{Estimate} & \textbf{Est.Error} & \textbf{95\% CI} & \textbf{Rhat} & \textbf{Bulk\_ESS} & \textbf{Tail\_ESS} \\
\hline
Intercept{[1]} & $-1.66$ & $0.18$ & $[-2.03, -1.30]$ & $1.00$ & $1555$ & $2714$ \\
Intercept{[2]} & $-0.10$ & $0.18$ & $[-0.46,\ 0.25]$ & $1.00$ & $1459$ & $2539$ \\
Intercept{[3]} & $0.95$  & $0.18$ & $[0.58,\ 1.31]$  & $1.00$ & $1433$ & $2499$ \\
Intercept{[4]} & $1.38$  & $0.18$ & $[1.02,\ 1.74]$  & $1.00$ & $1428$ & $2398$ \\
Intercept{[5]} & $2.36$  & $0.19$ & $[1.99,\ 2.74]$  & $1.00$ & $1489$ & $2857$ \\
Intercept{[6]} & $3.71$  & $0.21$ & $[3.29,\ 4.13]$  & $1.00$ & $1664$ & $3693$ \\
\hline
Determiner (few vs many)      & $0.06$  & $0.11$ & $[-0.16,\ 0.29]$ & $1.00$ & $4381$ & $5275$ \\
Grammaticality (illusion)     & $-0.19$ & $0.11$ & $[-0.41,\ 0.03]$ & $1.00$ & $4566$ & $5845$ \\
Grammaticality (grammatical)  & $1.51$  & $0.16$ & $[1.20,\ 1.82]$  & $1.00$ & $3405$ & $4449$ \\
\hline
Determiner $\times$ Grammaticality (illusion)     & $0.58$  & $0.14$ & $[0.31,\ 0.86]$  & $1.00$ & $6385$ & $5152$ \\
Determiner $\times$ Grammaticality (grammatical)  & $-0.64$ & $0.15$ & $[-0.95, -0.35]$ & $1.00$ & $5368$ & $5410$ \\
\hline
\end{tabular}
\label{tab:s_fewmany_model_result}
\end{table}

\begin{table}[ht]
\caption{Statistical results for the model in Experiment 2 (\textit{few}, \textit{most})}
\centering
\begin{tabular}{p{3.7cm}lllllll}
\hline
\textbf{Term} & \textbf{Estimate} & \textbf{Est.\ Error} & \textbf{95\% CI} & \textbf{Rhat} & \textbf{Bulk\_ESS} & \textbf{Tail\_ESS} \\
\hline
Intercept{[1]} & $-2.02$ & $0.17$ & $[-2.37, -1.69]$ & $1.0$ & $2185$ & $4050$ \\
Intercept{[2]} & $-0.20$ & $0.16$ & $[-0.52,\ 0.12]$ & $1.0$ & $1902$ & $3148$ \\
Intercept{[3]} & $0.94$  & $0.16$ & $[0.62,\ 1.26]$  & $1.0$ & $2047$ & $3212$ \\
Intercept{[4]} & $1.34$  & $0.16$ & $[1.01,\ 1.66]$  & $1.0$ & $2081$ & $3500$ \\
Intercept{[5]} & $2.35$  & $0.17$ & $[2.02,\ 2.69]$  & $1.0$ & $2200$ & $4229$ \\
Intercept{[6]} & $3.57$  & $0.19$ & $[3.20,\ 3.96]$  & $1.0$ & $2961$ & $4761$ \\ \hline
determiner (few\ldots most)      & $0.09$  & $0.11$ & $[-0.13,\ 0.31]$ & $1.0$ & $7824$  & $6520$ \\
grammaticality (illusion)        & $-0.22$ & $0.10$ & $[-0.41, -0.02]$ & $1.0$ & $10391$ & $6613$ \\
grammaticality (grammatical)               & $1.37$  & $0.17$ & $[1.04,\ 1.71]$  & $1.0$ & $6462$ & $5823$ \\ \hline
few\ldots most $\times$ illusion           & $0.46$  & $0.14$ & $[0.17,\ 0.74]$  & $1.0$ & $8399$ & $5963$ \\
few\ldots most $\times$ grammatical        & $-0.51$ & $0.13$ & $[-0.77, -0.24]$ & $1.0$ & $9043$ & $7157$ \\
\hline
\end{tabular}
\label{tab:s_fewmost_model_result}
\end{table}

\begin{table}[ht]
\caption{Statistical results for the model in Experiment 3 (\textit{few}, \textit{all})}
\centering
\begin{tabular}{p{3.7cm}lllllll}
\hline
\textbf{Term} & \textbf{Estimate} & \textbf{Est.\ Error} & \textbf{95\% CI} & \textbf{Rhat} & \textbf{Bulk\_ESS} & \textbf{Tail\_ESS} \\
\hline
Intercept{[1]} & $-1.61$ & $0.19$ & $[-1.98, -1.25]$ & $1.01$ & $793$  & $2147$ \\
Intercept{[2]} & $-0.04$ & $0.19$ & $[-0.40,\ 0.31]$ & $1.01$ & $770$  & $1814$ \\
Intercept{[3]} & $1.01$  & $0.19$ & $[0.65,\ 1.36]$  & $1.01$ & $776$  & $1849$ \\
Intercept{[4]} & $1.49$  & $0.19$ & $[1.13,\ 1.85]$  & $1.01$ & $785$  & $1863$ \\
Intercept{[5]} & $2.40$  & $0.19$ & $[2.02,\ 2.76]$  & $1.01$ & $829$  & $2110$ \\
Intercept{[6]} & $3.71$  & $0.22$ & $[3.28,\ 4.14]$  & $1.0$  & $1091$ & $2742$ \\ \hline
determiner (few\ldots all)           & $-0.38$ & $0.11$ & $[-0.59, -0.17]$ & $1.0$ & $2843$ & $4584$ \\
grammaticality (illusion)            & $-0.26$ & $0.10$ & $[-0.45, -0.07]$ & $1.0$ & $4256$ & $5192$ \\
grammaticality (grammatical)         & $1.08$  & $0.16$ & $[0.77,\ 1.39]$  & $1.0$ & $2280$ & $4177$ \\ \hline
few\ldots all $\times$ illusion      & $0.47$  & $0.14$ & $[0.21,\ 0.75]$  & $1.0$ & $4274$ & $5042$ \\
few\ldots all $\times$ grammatical   & $-0.09$ & $0.14$ & $[-0.36,\ 0.18]$ & $1.0$ & $3910$ & $5144$ \\
\hline
\end{tabular}
\label{tab:s_fewall_model_result}
\end{table}

\begin{table}[ht]
\caption{Statistical results for the model in Experiment 4 (\textit{few},\textit{the})}
\centering
\begin{tabular}{p{3.7cm}lllllll}
\hline
\textbf{Term} & \textbf{Estimate} & \textbf{Est.\ Error} & \textbf{95\% CI} & \textbf{Rhat} & \textbf{Bulk\_ESS} & \textbf{Tail\_ESS} \\
\hline
Intercept{[1]} & $-2.13$ & $0.17$ & $[-2.48, -1.79]$ & $1.01$ & $1135$ & $2733$ \\
Intercept{[2]} & $-0.37$ & $0.16$ & $[-0.69, -0.06]$ & $1.01$ & $1028$ & $2519$ \\
Intercept{[3]} & $0.57$  & $0.16$ & $[0.25,\ 0.89]$  & $1.01$ & $1045$ & $2299$ \\
Intercept{[4]} & $1.09$  & $0.17$ & $[0.76,\ 1.41]$  & $1.01$ & $1072$ & $2465$ \\
Intercept{[5]} & $2.01$  & $0.17$ & $[1.67,\ 2.33]$  & $1.01$ & $1117$ & $2621$ \\
Intercept{[6]} & $3.41$  & $0.18$ & $[3.05,\ 3.77]$  & $1.0$  & $1354$ & $3010$ \\ \hline
no\ldots the $\times$ grammatical  & $1.37$ & $0.16$ & $[1.06,\ 1.67]$ & $1.0$ & $1957$ & $3478$ \\
few\ldots the $\times$ grammatical & $1.45$ & $0.16$ & $[1.12,\ 1.77]$ & $1.0$ & $1942$ & $2821$ \\
no\ldots the $\times$ illusion     & $-0.06$ & $0.11$ & $[-0.27,\ 0.15]$ & $1.0$ & $4655$ & $4875$ \\
few\ldots the $\times$ illusion    & $0.12$ & $0.12$ & $[-0.11,\ 0.35]$ & $1.0$ & $4131$ & $5455$ \\
\hline
\end{tabular}
\label{tab:s_fewthe_model_result}
\end{table}

\begin{table}[ht]
\caption{Statistical results for the model in Experiment 5 (\textit{no}, \textit{many})}
\centering
\begin{tabular}{p{3.7cm}lllllll}
\hline
\textbf{Term} & \textbf{Estimate} & \textbf{Est.\ Error} & \textbf{95\% CI} & \textbf{Rhat} & \textbf{Bulk\_ESS} & \textbf{Tail\_ESS} \\
\hline
Intercept{[1]} & $-1.40$ & $0.18$ & $[-1.76, -1.04]$ & $1.0$ & $1797$ & $3317$ \\
Intercept{[2]} & $0.17$  & $0.18$ & $[-0.19,\ 0.52]$ & $1.0$ & $1732$ & $2914$ \\
Intercept{[3]} & $0.99$  & $0.18$ & $[0.63,\ 1.34]$  & $1.0$ & $1761$ & $2946$ \\
Intercept{[4]} & $1.37$  & $0.18$ & $[1.02,\ 1.73]$  & $1.0$ & $1778$ & $3014$ \\
Intercept{[5]} & $2.20$  & $0.19$ & $[1.84,\ 2.57]$  & $1.0$ & $1854$ & $3161$ \\
Intercept{[6]} & $3.35$  & $0.20$ & $[2.95,\ 3.75]$  & $1.0$ & $2098$ & $3432$ \\ \hline
determiner (no\ldots many)           & $0.18$  & $0.10$ & $[-0.03,\ 0.38]$ & $1.0$ & $7108$ & $5930$ \\
grammaticality (illusion)            & $0.11$  & $0.12$ & $[-0.12,\ 0.33]$ & $1.0$ & $7909$ & $6057$ \\
grammaticality (grammatical)         & $1.38$  & $0.19$ & $[1.01,\ 1.75]$  & $1.0$ & $4216$ & $4846$ \\ \hline
no\ldots many $\times$ illusion      & $-0.09$ & $0.14$ & $[-0.37,\ 0.19]$ & $1.0$ & $8391$ & $6397$ \\
no\ldots many $\times$ grammatical   & $-0.81$ & $0.15$ & $[-1.10, -0.51]$ & $1.0$ & $7338$ & $6491$ \\
\hline
\end{tabular}
\label{tab:s_nomany_stats}
\end{table}

\begin{table}[t]
\caption{Statistical results for the model with interaction in Experiment 6 (\textit{no}, \textit{all})}
\centering
\begin{tabular}{p{3.7cm}lllllll}
\hline
\textbf{Term} & \textbf{Estimate} & \textbf{Est.\ Error} & \textbf{95\% CI} & \textbf{Rhat} & \textbf{Bulk\_ESS} & \textbf{Tail\_ESS} \\
\hline
Intercept{[1]} & $-1.77$ & $0.17$ & $[-2.11, -1.45]$ & $1.01$ & $979$  & $2412$ \\
Intercept{[2]} & $-0.09$ & $0.16$ & $[-0.42,\ 0.22]$ & $1.01$ & $900$  & $1893$ \\
Intercept{[3]} & $1.00$  & $0.17$ & $[0.67,\ 1.32]$  & $1.01$ & $919$  & $2269$ \\
Intercept{[4]} & $1.26$  & $0.17$ & $[0.92,\ 1.58]$  & $1.01$ & $932$  & $2221$ \\
Intercept{[5]} & $2.24$  & $0.17$ & $[1.90,\ 2.58]$  & $1.01$ & $995$  & $2313$ \\
Intercept{[6]} & $3.21$  & $0.19$ & $[2.83,\ 3.59]$  & $1.01$ & $1155$ & $2478$ \\ \hline
determiner (no\ldots all)            & $-0.25$ & $0.10$ & $[-0.45, -0.05]$ & $1.0$ & $3878$ & $5295$ \\
grammaticality (illusion)            & $-0.30$ & $0.10$ & $[-0.49, -0.11]$ & $1.0$ & $4581$ & $5641$ \\
grammaticality (grammatical)         & $1.17$  & $0.14$ & $[0.90,\ 1.46]$  & $1.0$ & $2990$ & $4513$ \\ \hline
no\ldots all $\times$ illusion       & $0.32$  & $0.14$ & $[0.04,\ 0.60]$  & $1.0$ & $4181$ & $5442$ \\
no\ldots all $\times$ grammatical    & $-0.53$ & $0.15$ & $[-0.84, -0.23]$ & $1.0$ & $3990$ & $5127$ \\
\hline
\end{tabular}
\label{tab:s_noall_model_stats}
\end{table}

\end{document}